# Tackling Multimodal Device Distributions in Inverse Photonic Design using Invertible Neural Networks

*Michel Frising[†,*], Jorge Bravo-Abad[‡], Ferry Prins[†]*

AUTHOR ADDRESS

[†]Condensed Matter Physics Center (IFIMAC) and Department of Condensed Matter Physics, Autonomous University of Madrid, 28049 Madrid, Spain

[‡]Condensed Matter Physics Center (IFIMAC) and Department of Theoretical Condensed Matter Physics, Autonomous University of Madrid, 28049 Madrid, Spain

ABSTRACT: Inverse design, the process of matching a device or process parameters to exhibit a desired performance, is applied in many disciplines ranging from material design over chemical processes and to engineering. Machine learning has emerged as a promising approach to overcome current limitations imposed by the dimensionality of the parameter space and multimodal parameter distributions. Most traditional optimization routines assume an invertible one-to-one mapping between the design parameters and the target performance. However, comparable or even identical performance may be realized by different designs, yielding a multimodal distribution of possible solutions to the inverse design problem which confuses the optimization algorithm. Here, we show how a generative modeling approach based on invertible neural networks can provide the full distribution of possible solutions to the inverse design problem and resolve the ambiguity of nanodevice inverse design problems featuring multimodal distributions. We implement a Conditional Invertible Neural Network (cINN) and apply it to a proof-of-principle nanophotonic problem, consisting in tailoring the transmission spectrum of a metallic film milled by subwavelength indentations. We compare our approach with the commonly used conditional Variational Autoencoder (cVAE) framework and show the superior flexibility and accuracy of the proposed cINNs when dealing with multimodal device distributions. Our work shows that invertible neural networks provide a valuable and versatile toolkit for advancing inverse design in nanoscience and nanotechnology.

Inverse design is the process of matching a device or process parameters to produce a desired performance. The possibility of enabling new materials and nanodevices by 'reverse-engineering' them from the desired properties and characteristics has drawn a great deal of both fundamental and applied interest. Inverse design is particularly popular in the field of Nanophotonics, where the ongoing quest for miniaturization requires the exploration of extremely large parameter spaces spanned by freeform geometries and a wide variety of materials combinations[1]. To efficiently explore these vast parameter spaces, the inverse-design problem is commonly approached using computational optimization techniques capable of identifying solutions beyond human intuition. Techniques such as gradient-based topology optimization[2–6], evolutionary design[7,8] and more recently Artificial Neural Networks[9–12] and Global Optimization Nets[13,14] have been used successfully to design devices that vastly outperform designs based on human intuition.

To date, however, most inverse design approaches rely on the assumption that a unique one-to-one mapping exists between the device and a given design target. In reality, it is often the case that multiple device designs exhibit comparable or even identical performance, yielding a multimodal device distribution. One clear example of this scenario occurs when considering the symmetries of a system or a device. For example, the simple slit flanked by –arrays of grooves structure in Fig. 1 (which we will use throughout this manuscript for illustrative purposes) consists of a central slit in a thin metal film flanked on each side by different gratings consisting of periodic indentations in the metal. The system has a natural plane of symmetry in the middle of the slit as it does not matter for the spectral response if the gratings are swapped. While this may be obvious to the reader, this situation is less clear to an optimization routine which tries to minimize a target and assumes that there is, at least locally, a minimum. Consequently, the optimization routine may oscillate between the two solutions, preventing the algorithm from converging.

Several studies have attempted to address the issue of device distributions, though with limited success. Liu et al.[10] observed the problem of multimodality when trying to teach a neural network to reverse-engineer stacks of dielectrics to exhibit desired transmission spectrum. While the forward simulation predicting the transmission spectrum from device parameters could be trained with ease, they observed that matching a stack to a given spectrum (i.e. the inverse pass) was impossible. They attributed this problem to multimodality, meaning that different dielectric stacks map to the same response, preventing the network from converging. They circumvented this issue by using a tandem network in which the forward and backward passes feed into each other to stabilize training. While this approach allows the network to converge despite the multimodality of the training set, the trained network only offers a single solution while other possible solutions are lost.

To truly move beyond one-to-one mapping, recent approaches have explored the use of generative models, including for example Variational Autoencoders (VAEs)[15] and Generative Adversarial Networks (GANs)[11,16]. Rather than assuming a simple one-to-one mapping, these generative methods model a distribution of possible devices and their design parameters allowing for the identification of multiple solutions. Ma et al[17]. employed such a generative model for inverse photonic design, training a Variational Autoencoder (VAE) to generate proposals for a unit cell of a periodic metamaterial exhibiting a set of desired transmission and reflection properties. However, VAEs are restricted to simple parametrized distributions in their latent space, limiting their expressive power when the device distribution is multimodal. Normalizing flows have emerged as a powerful tool to construct more expressive distributions[18–22] beyond Gaussians to model the true data distribution, but have so far not been used in inverse photonic design. Conditional generative adversarial networks cGANS[23] are also in principle capable of modeling multimodal distributions of parameters but are very prone to mode collapse[24] meaning that the multimodal data distribution is mapped to only one of the modes while still ignoring other possible solutions.

Here, we propose the use of conditional Invertible Neural Networks (cINNs)[25,26] to tackle the issue of multimodal device parameters in inverse photonic design. cINNs have been recently introduced as a highly versatile platform for inverse design. They belong to the family of flow-based techniques[25,27], meaning they are capable of expressing complicated distributions. Since they are trained with maximum likelihood loss, mode collapse as observed in GANs, is virtually impossible[25]. Moreover, since the same network is used for the forward and backward pass, only half of number of parameters is necessary. cINNs have been shown to work in a wide range of applications, from improving the robustness in medical imaging and inverse kinematics[25] to changing the style of images[28]. Here, we apply cINNs to the field of photonics, demonstrating their capability to effectively deal with multimodal device parameters in inverse photonic design. As an example of our approach, we train the cINN to find the geometrical parameters of the aforementioned slit in a thin metal film flanked by periodic grooves[29,30] to match a desired transmission spectrum. The slit flanked by gratings is particularly suited for this purpose, as the symmetry of the structure intrinsically introduces a multimodal distribution in the parameter space of the device geometry. In contrast to past approaches, we show that the cINN provides all possible solutions to the inverse design problem. We emphasize this by comparing our results to a VAE network trained on the same data.

The structures under study are shown schematically shown in Figure 1. A central subwavelength aperture of width $a_0$ in a silver film of thickness $t$ is surrounded on both sides by gratings made from a finite periodic array of grooves. The periodicity $\Lambda$ and height $h$ of the grooves can be different on each side of the slit. The structure is completely described by the parameter vector $x = [\Lambda_1, \Lambda_2, h_1, h_2]$. The width of the groves is fixed to always be half of the corresponding periodicity, i.e. a duty cycle of 50 %, and the entire structure is assumed to be embedded in air with a refractive index n=1.

Extraordinary optical transmission occurs when the incident light (in the form of surface plasmon polaritons, or SPPs) constructively interferes at the subwavelength aperture. By introducing variations in the geometry of the slit array, the wavelengths at which constructive interference occurs can be modulated, allowing for control over the transmission spectrum of the structure. For a slit which is flanked by the same gratings on each side (i.e. $\Lambda_1 = \Lambda_2$ and $h_1 = h_2$), a single transmission peak dominates the transmission spectrum. For an asymmetric configuration, i.e. $\Lambda_1 \neq \Lambda_2$ and $h_1 \neq h_2$, more complex transmission spectra can occur, for example yielding two separate transmission peaks at different wavelengths, each corresponding to one of the two periodicities of the gratings. Intuitively, the mirror image of each asymmetric configuration has an identical transmission spectrum, illustrating the intrinsic multimodality that is present as a result of symmetries in the geometry. Importantly, while this type of multimodality of mirrored designs is straightforward for human intuition, it causes severe problems in inverse photonic design where one target spectrum suddenly has multiple design solutions associated with it.

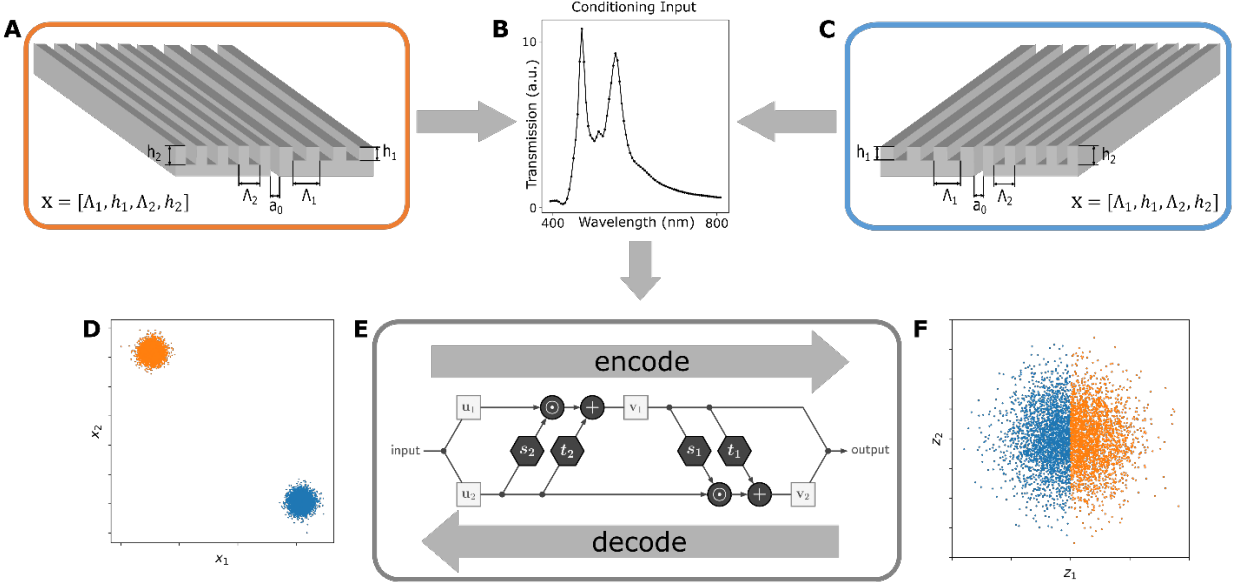

**Figure 1** A and C show the devices under investigation: a slit with fixed width $a_0$ flanked by gratings on each side with periodicity $\Lambda_1$ and $\Lambda_2$, respectively, and a duty cycle of 50%. The two devices in A and C are mirrored along a plane going through the center of the slit and have the same transmission spectrum as shown in B. The cINN in E takes the simulated spectrum corresponding to a specific device and maps the device parameters in D to the latent space variable z which is a Gaussian in F. Since two two different devices generate the same transmission spectrum the device parameter space in D show to modes. The cINN is capable of mapping this complex distribution to the simple Gaussian in F that can be easily sampled.

The transmission through the slit flanked by gratings is simulated using a Coupled Mode Theory (CMT) framework[30] The CMT framework has been used extensively to simulate transmission and EOT phenomena in a variety of systems[29,30] and is known to provide accurate results with minimal computational cost. As a result, the generation of a full training set of 60,000 different structures can be performed efficiently. The training set is divided between 45,000 structures for training, and 15,000 structures for validation.

We modeled our cINN after Ardizzone et al.[26] using their FrEIA framework, the technical details are described in the Methods section and the Supplementary Information. Briefly, the cINN takes the device parameter vector and maps it to a latent space variable *z*, which can be sampled conveniently to generate new devices by running the cINN in reverse. The cINN additionally takes a conditional vector extracted from the spectrum corresponding to the device as an input. The conditioning network consists of a ResNet-34[31]. To benchmark the performance of our cINN, we compare it to a commonly used cVAE[32].

**Results and Discussion**

The power and flexibility of the cINN for this class of problems is showcased in Figure 2 Two target spectra from the validation set are chosen (black solid lines in A and B). Each time, $10^4$ device parameters are generated by sampling from the trained cINN with the spectra in A and B as conditioning inputs. The resulting distributions of parameters of the devices exhibiting these transmission spectra are shown as histograms in C-F and G-J. The solid black line with the triangle on top marks the original parameter vector that has been used in the simulation. The generated parameters are then passed to a forward network which has been trained previously. The forward network is a fully connected dense network which takes device parameters as input and outputs the transmission spectrum of that device and its sole purpose is to speed up the generation process. In the Supporting Information evidence is presented that this forward network indeed models the generative process with high fidelity. The regenerated spectra show excellent agreement with the target spectrum. The mean of the $10^4$ regenerated spectra fall exactly on top of the target and the shaded regions present the $2^{nd}$ to $98^{th}$ percentile region, meaning that 96% of the regenerated spectra lie within that region. The cINN correctly learns that the spectrum in A has been generated from gratings with the same periodicity on both sides of the slit (panels C and E in Fig. 2), while the spectrum in B requires two different periodicities (panels G and I in Fig. 2). Also note that, as previously mentioned, the dataset was constructed randomly and the network was still able to fully capture the underlying symmetry properties of the problem. It is also interesing to note that in panels D and F the network generated two connected peaks for the groove depth. Since the regenerated spectra show excellent agreement it seems that the network is not so sensitive to groove depth for that specific conditioning input. In panels H and G of Fig. 2, the proposed grove depths are clearly separated.

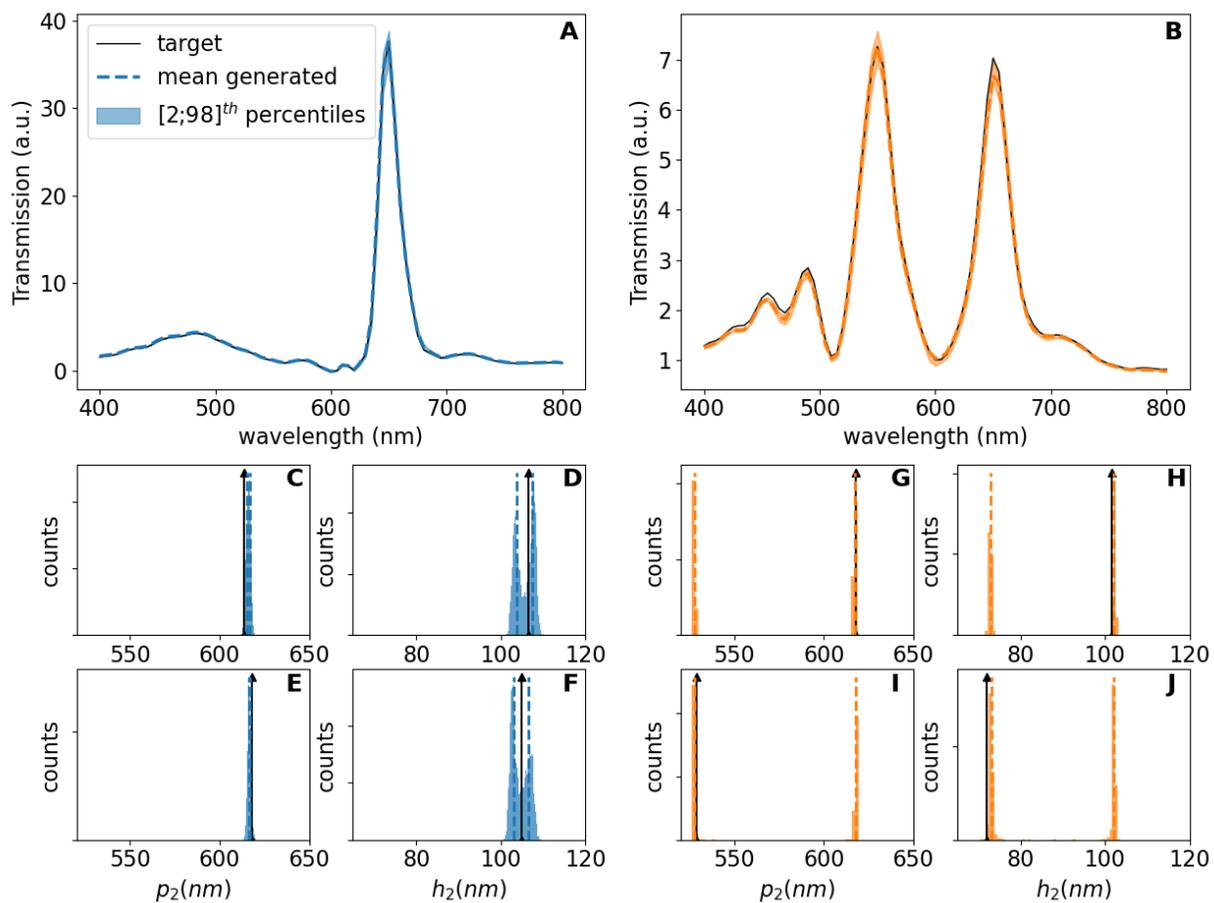

Figure 2 A and B: Two randomly chosen spectra from the validation set the network has never seen before, shown as solid black lines. C-F, G-J: The generated parameters conditioned on that specific spectrum. Each time $10^4$ have been sampled. The solid line with the triangle on top is the original parameter vector used in the simulations. The generated parameters and then used to reconstruct the corresponding spectra with the forward network. The mean of the generated spectra is shown as a dashed line in A-B and perfectly agree with the conditioning input. The shaded regions in A and B are the [2,98] percentile intervals showing how confident the network is about the solution.

The added flexibility of the normalizing flow is apparent when comparing the latent space generated by the cINN with the latent space of a cVAE[32 17] in Fig. 3. In our example structure, the latent space has dimensionality 4 and we chose one spectrum as conditioning input and sampled 10,000 samples as before. Figure 3 shows scatter plots of generated periodicities $p_1$ and $p_2$ for the cINN (Fig. 3A) and the cVAE (Fig 3B). In the case of the cINN, the vast majority of solutions concentrates around the target values for $p_1$ and $p_2$, demonstrating the excellent performance of this network. Please note that the two clusters are connected by a small number of points due to the fact that the specific normalizing flow that we used maintains continuity and cannot split a unimodal base distribution. The result, however, is a good enough approximation, since these points correspond to 400 samples out of $10^4$ or 4%. The cVAE, however, has a much larger spread in the latent space, showing both less precise and less accurate performance as compared to the cINN.

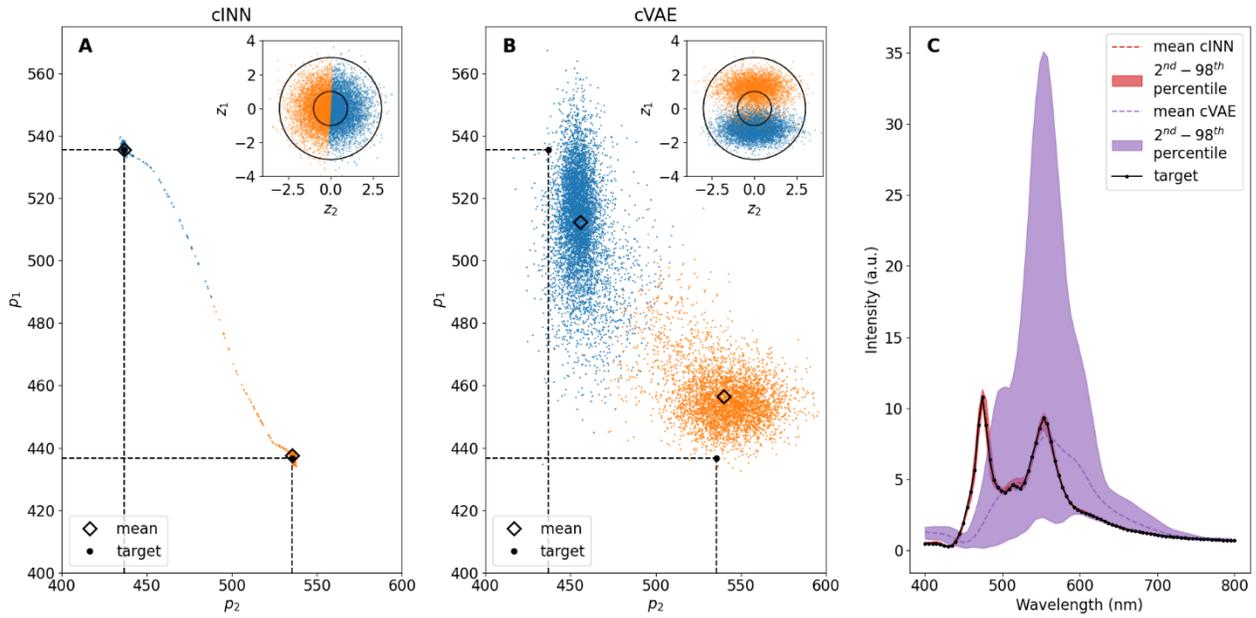

Figure 3 A Given the conditioning input shown in C as the black solid line, 10'000 device parameters are generated with the cINN. The true device parameters are indicated by a x, the mean of the generated clusters as diamonds. The clusters were separated with k-means clustering. The cINN learns to home in on the true device parameters that give rise to this particular spectrum. B 10'000 device parameters generated with the cVAE. Again, the means of the generated parameters are indicated by diamonds and it is clear that the cVAE is struggling to focus. When looking at the latent space produced by the cVAE in the inset, the problem becomes apparent: the latent space is not well approximated by a Gaussian. The latent space of the cINN, however, is well approximated by a Gaussian and can therefore be easliy sampled. Note that the two clusters in A are connected by a line which is due to the fact that the normalizing flow of the cINN can only deform the Gaussian latent space to match the data distribution, mantaining continuity.

To understand the large difference in the performance of the cINN and the cVAE, we can have a closer look at their respective latent spaces shown as the insets of Fig. 3 A and B. The complete latent space is reproduced in the Supporting Information, here only $z_1$ and $z_2$ are shown since they capture the relevant information. While the latent space of the cINN for the two solutions is a Gaussian, the cVAE is far from normally distributed. The resulting limitation for sampling becomes obvious: With the cINN we can directly sample from a Gaussian with zero mean and a variance of one in the latent space and obtain accurate results from the inverse pass of the cINN. If samples are drawn with the same method from the latent space of the cVAE, a large number of out-of-distribution samples are generated, resulting in incorrect predictions. In both cases the generated parameter vectors were used to re-simulate the transmission spectra of the proposed devices as shown in Fig. 3C. The cINN shows excellent agreement as before, while the results from the cVAE are neither accurate nor precise. Please note there are techniques to alleviate this problem by sampling from the aggregated posterior as suggested by Tomczak and Welling[33], or by generating another VAE for sampling as suggested by Dai and Wipf[34], but that would add an additional layer of complexity. As mentioned before, normalizing flows as described in Kingma and al.[18] would theoretically allow to model more flexible posterior distributions but add more complexity as well. In contrast, the cINN offers a powerful yet simple framework to model complex data distributions.

At this point it is important to highlight how the model assumptions and loss functions are hindering training on multimodal distributions or even leading it to failure. First, consider the commonly used mean squared error (MSE) which is used in a regression setting. If the phenomenon to be modeled is strictly monotonic, there is only one solution and the MSE is a good measure for the fit, since the solution that best fits the data will also be the one that is closest to the data point. However, if the data is multimodal, a point at equal distance between two or more solutions is the one minimizing the MSE, but the proposed solution is very off from any of the actual solutions. As discussed earlier, a cVAE struggles with multimodal datasets because the Gaussian for the posterior, while being easy to sample, is very restrictive and the model ends up lacking expressive power[35] or the sampling becomes more tedious. And while cVAE can be augmented with normalizing flows to model multimodal distributions, this addition makes them hard to train. Using a mixture of Gaussians[36] for the latent space or a Vamp Prior[33] allows to build more expressive posteriors but also sacrifices simplicity and ease to train. The cINN, by construction, leads itself to the modeling of complicated data distributions and no assumptions need to be made about the posterior.

**Conclusion**

In summary, we have shown how a multimodal device distribution can lead to pitfalls with commonly used generative models to learn these device distributions. Furthermore, we demonstrated the flexibility of cINNs how they can, with no additional knowledge about the device distribution, learn a mapping that can be used to sample new structures, providing the full posterior of the device distribution, meaning all the possible solutions to the inverse design problem. In general, adopting a probabilistic approach provides a more complete picture of all the possible solutions to an inverse design problem and how confident the algorithm is. Here we note that very recently the use of cINNs has been mentioned in the context of benchmarking different deep learning approaches to inverse models for designing artificial electromagnetic materials[37], but owithout considering multimodal device distributions. On a similar note, GANs have also been employed for inverse photonic design[11,16] but have not been thoroughly explored to generate distributions of devices in the context of generative modeling.

The cINN provides solutions with high accuracy and precision on all solutions in the design space, whereas the cVAE only captures the fact that the solution space is multimodal, but with low precision and accuracy. The reason for that is the limited flexibility unimodal Gaussian latent variables provide. While flow-based models and more expressive priors could alleviate that problem, the cINN offers a simple and straightforward framework and simple to train solution. Finally, it is important to emphasize that while we have focussed on the sslit array as a proof-of principle problem the advantage of the cINN in solving multimodal device distributions is generic and may help solve a large variety of inverse design problems to explore the design space of nanophotonic devices.

**Methods**

We implemented a cINN following the approach of Ardizzone et al., using their FrEIA framework.[26] As shown schematically in **Figure 1**, the basic building block of the cINN is the (conditional) affine coupling block first proposed by Dinh et al.[27]. The network models a change of variables that maps a latent variable z to a sample x with the conditional input c:

$$p_x(f(z,c)) = p_z(z) \left|\det \frac{\partial f(z,c)}{\partial z}\right|^{-1} \qquad (1)$$

The conditioning input, in this case the spectrum $y$ that corresponds to a certain device parameter vector $x$, is fed through an additional conditioning network consisting of a ResNet-34[31] to extract meaningful features from the spectrum. More details for the hyperparameters and data preparation of the network can be found in the supporting information.

The cVAE used for comparison is implemented similarly to Ma et al.[17] and Sohn et al.[32] and consists of two simple dense networks with 5 hidden layers and 256 neurons each. The conditioning input is concatenated to the input of the encoder and decoder, respectively. One network maps the device parameter along with the simulated spectrum to the latent variable $z$ (encoder), while the other network tries to reconstruct the parameter vector (decoder) from the latent variable $z$ and the simulated spectrum.

The main difference between the two networks resides in the way they are trained. The cINN is trained with a modified maximum likelihood loss function including the Jacobian from the change of variables as suggested by Kruse et al.[38]:

$$\mathcal{L}(z) = \frac{z^2}{2} - log|det J_{x \to z}| \qquad (2)$$

while the cVAE is trained with the evidence lower bound (ELBO)[32]:

$$\mathcal{L}(z, y) = -D_{KL}\big(q_\theta(z|x_i, y_i)||p(z|y)\big) + E_{z \sim q_\theta(z|x_i, y_i)}[log\, p_\phi(x_i|z, y_i)] \qquad (3)$$

where $D_{KL}$ is the Kullback-Leibler divergence, which measures the difference between distributions $q_\theta(z|x_i, y_i)$ and $p(z|y)$, the approximated posterior and the prior, respectively. The dependence on $y$ in the prior $p(z|y)$ has been dropped to simplify training as suggested by Sohn et al[32]. $E_{z \sim q_\theta(z|x_i, y_i)}[log\, p_\phi(x_i|z, y_i)]$ is the reconstruction loss. under the constraint that the latent space should be Gaussian, meaning that the term $D_{KL}\big(q_\theta(z|x_i, y_i)||p(z|y)\big)$ acts as a regularization term[39,40]. If $q_\theta(z|x_i, y_i)$ is a Gaussian with mean $\mu_q$ and variance $\sigma_q$ and $p(z|y)$ a Gaussian with mean $\mu_p$ and variance $\sigma_p$, the Kullback-Liebler divergence simplifies to its well known form[15,40]:

$$-D_{KL}\big(q_\theta(z|x_i, y_i)||p(z|y)\big) = \frac{1}{2}[1 + \log \sigma_q^2 - \sigma_q^2 - \mu_q^2] \qquad (4)$$

Where $\mu_q$ and variance $\sigma_q$ are estimated by the encoder network.

The loss functions in Eqs. (2) and (3) are very different, reflecting the different objectives to minimize. While the goal of Eq. (2) is to maximize the likelihood after the change of variables, Eq. (3) tries to minimize reconstruction loss while imposing the restriction of a normal distribution on the latent space, like a regularization term[39]. Since Eq. 2 is a form of maximum likelihood loss function, mode collapse is virtually impossible[28]. The objective of the ELBO is to learn a representation of the data that is as close as possible to the true distribution which is especially useful when the data is very high dimensional as normalizing flows typically do not scale well to high dimensional data such as images.

Even though we have access to relatively fast simulations to obtain the device response, we found it beneficial to train a simple fully connected model consisting of three hidden layers with 512 neurons each and relu activation for to predict the device response for a given parameter vector. This is not necessary but useful to quickly visualize the behavior of the devices generated by the cINN and the cVAE.

**Author Contributions**

The manuscript was written by MF with suggestions from FP and JBA. MF setup and trained the Neural Networks shown. JBA wrote the original codes to simulate the investigated devices and generated the training set based on device parameters selected by MF. FP provides helpful suggestions and discussions throughout the whole process.

**Funding Sources**

The project that gave rise to these results received the support of a fellowship from the "la Caixa" Foundation (ID 100010434). The fellowship code is LCF/BQ/DI18/11660037. This project has received funding from the European Union's Horizon 2020 research and innovation programme under the Marie Skłodowska-Curie grant agreement No. 713673. The work was further supported by the Spanish Ministry for Science, Innovation, and Universities through the Europa Excelencia program (EUR2019-103826).

**Acknowledgement**

We graciously acknowledge the help of Lynton Ardizzone with feedback and suggestions on implementing our own cINN and helpful discussions to understand the INN framework. Furthermore, we want to thank Pablo Sanchez-Martin for helpful discussions about VAEs in general.

**Supporting Information**

The forward network

A simple dense network consisting of 5 dense layers with 512 neurons in each hidden layer and ReLU activation function has been trained to reproduce the spectral response of the device given the parameter vector describing it, replacing the CMT simulations. Figure S1 shows a histogram of the residues of the generated spectra from the test set, of which 99.9% are below 1.5%. To further illustrate the accuracy of the forward network, the generated spectra (orange) of 40 randomly chosen devices which are plotted together with the target (blue) and no deviations are apparent.

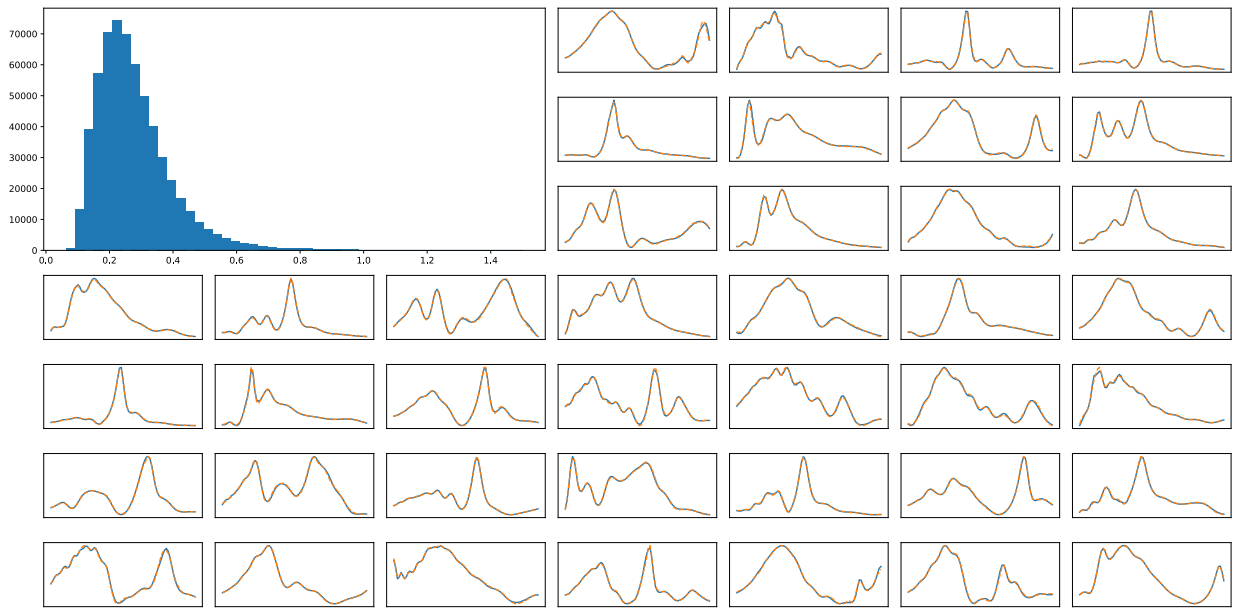

Figure S1 The histogram shows the residues of the simulated spectra and the spectra generated by the NN. 99.9% of the 600k samples have residues lower than 1.5 percent. The other panels show randomly selected spectra from the validation set to showcase the power of the forward simulating network.

Details on the cINN implementation

The cINN used 6 invertible all-in-one coupling blocks from the FrEIA framework (https://github.com/VLL-HD/FrEIA ). Each coupling block consists of a invertible affine coupling block as proposed by Dinh et al.[27] as shown in Fig. S2. The input $x$ is split into two vectors $x_1$ and $x_2$. The vector $x_2$ is passed through two neural networks $s$ and $t$. The outputs of these netwroks are then combined with $x_1$ such that $u_1 = x_1 \exp s(x_2) + t(x_2)$ and $u_2 = x_2$. This operation can be trivially inverted without the need to invert the NNs $s$ and $t$ to reconstruct x. The subnets $s$ and $t$ are chosen to be dense network with 1 hidden layer with 512 neurons and the whole cINN is made of 6 of these affine coupling blocks. Additionally the affine coupling block takes a conditional input which is the same for each coupling block. The conditioning input is provided by the conditioning network consisting of a ResNet-34[31] with a linear output layer of size 512 which is fed directly into the cINN. Also, we found that decomposing the spectra with a wavelet transformation before sending them through the conditioning network parameter distributions with lower variance. The network was trained for 300 epochs, batch size of 128 and a learning rate of $10^{-3}$ which was reduced to $10^{-4}$ after 200 epochs.

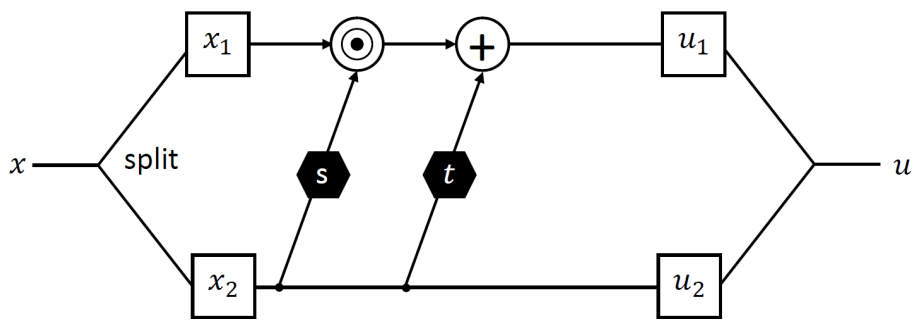

Figure S2 The affine coupling block from Dinh et al.[27]. The input $x$ is split into two vectors $x_1$ and $x_2$. The vector $x_2$ is passed through two neural networks $s$ and $t$. The outputs of these netwroks

are then combined with $x_1$ such that $u_1 = x_1 \exp s(x_2) + t(x_2)$ and $u_2 = x_2$. This operation can be trivially inverted without the need to invert the NNs $s$ and $t$ to reconstruct x.

The full latent spaces

Fig. S2 and S3 show the full latent space for the cVAE and the cINN, respectively. Since the latent space is four dimensional we choose to project the latent variables pairwise on the different axis. On the diagonal the distributions just along one dimension in latent space are shown. It is clearly visible in S2 how the latent space needs to be Gaussian and how two Gaussians need to be well separated in order for the modes not to have cross-talk which in turn is problematic for sampling. The cINN on the other hand can model arbitrary distributions such that the two modes can almost touch and the latent space more closely resembles a Gaussian.

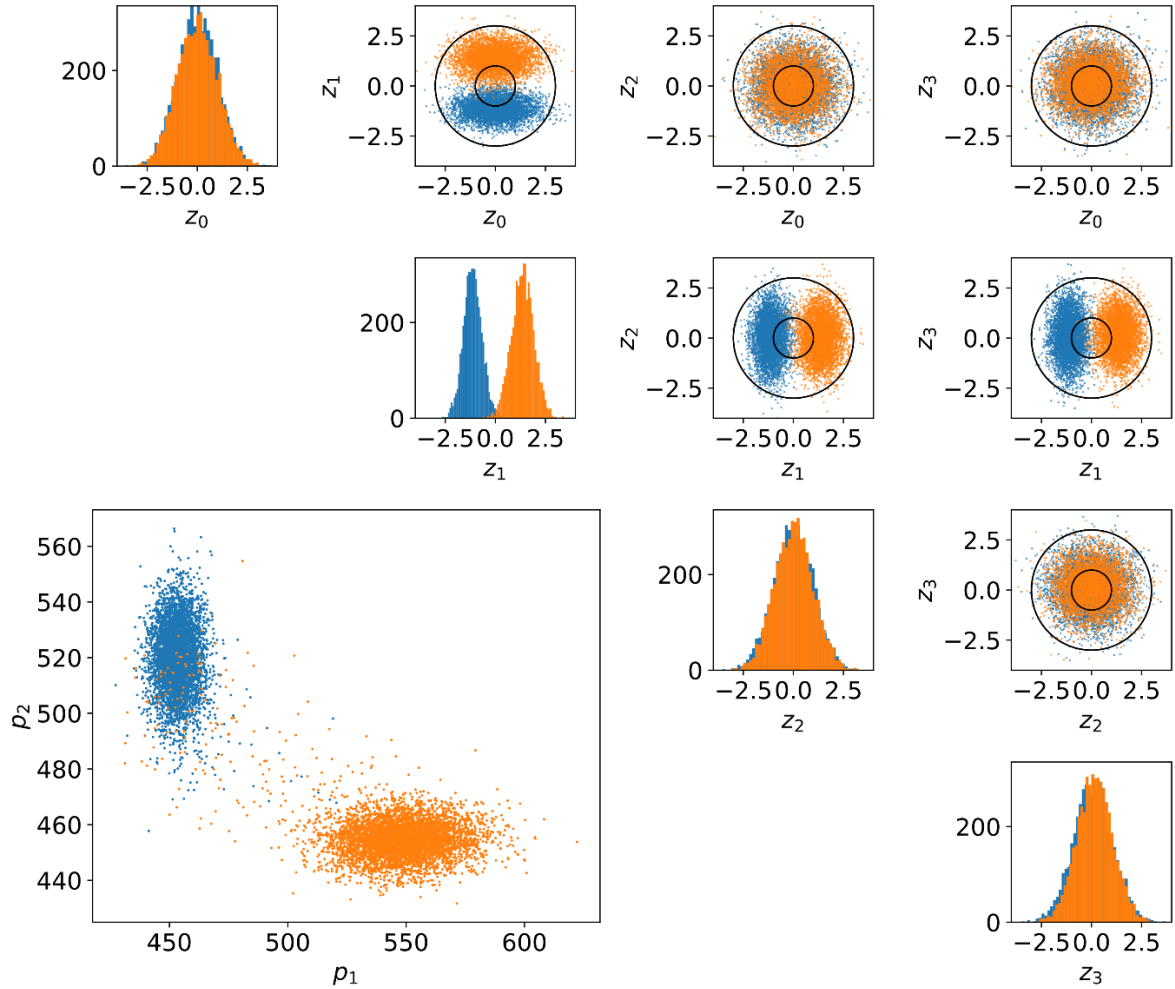

S2 The full latent space of the sample shown in the main text. The different axis are shown pairwise. Note how here $z_1$ divides the latent space in two different regions. Since the latent space is limited to be a Gaussian, its ability to represent a multimodal distribution is limited as only the non-overlapping regions of the two peaks can map to different modes.

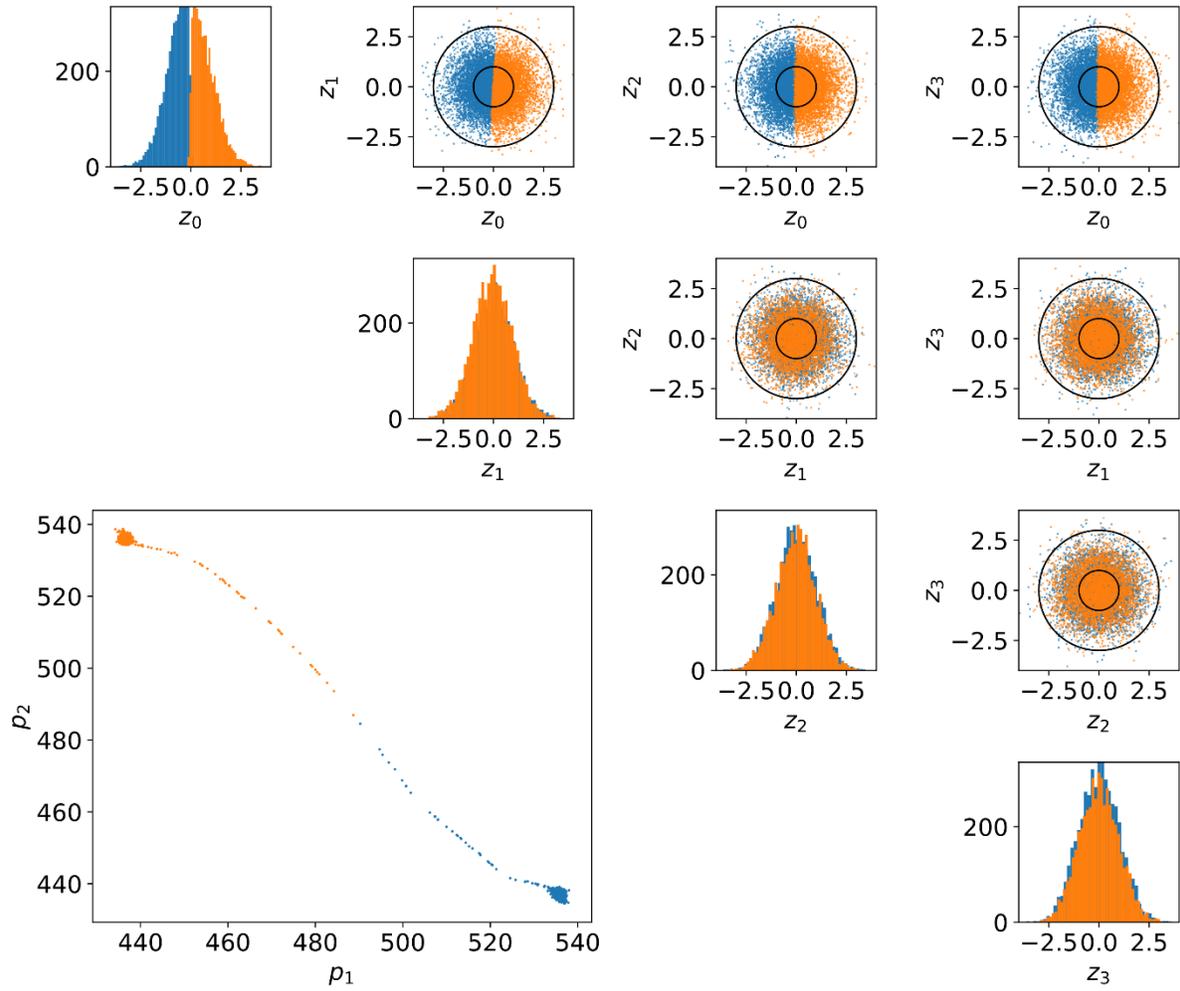

S2 The full latent space of the sample shown in the main text. The different axis are shown pairwise. Note how $z_0$ divides the latent space in two different regions.